\documentclass[10pt,twocolumn,letterpaper]{article}

\usepackage{wacv}
\usepackage{times}
\usepackage{epsfig}
\usepackage{graphicx}
\usepackage{amsmath}
\usepackage{amsthm}
\usepackage{amssymb}
\usepackage{bbm}
\usepackage{comment}
\usepackage{rotating}
\usepackage{color}
\newtheorem{thm}{Theorem}[section]

\newtheorem{prop}[thm]{Proposition}

\newcommand{\lclass}{\mathcal{L}_{\text{class}}}
\newcommand{\lneigh}{\mathcal{L}_{\text{neighb}}}

\newcommand{\faty}{\mathbf{Y}}

\newcommand{\fatxs}{\mathbf{x}}
\newcommand{\fatys}{\mathbf{y}}
\newcommand{\fatzs}{\mathbf{z}}

\newcommand{\x}{\times}

\newcommand{\prior}{p^{\text{prior}}}
\newcommand{\pcrf}{p^{\text{CRF}}}

\newcommand{\qfake}{q^{\text{aux}}}
\newcommand{\qmf}{q^{\text{MF}}}

\newcommand{\pnei}{p^{\text{neighb}}}

\newcommand{\KL}{\text{KL}}

\newcommand\MyIm[1]{%
  \includegraphics[width=2cm]{#1}
}

\wacvfinalcopy 

\def\wacvPaperID{150} 

\ifwacvfinal\fi
\begin{document}

\title{Learning to segment with image-level supervision}

\author{Gaurav Pandey \hspace{2cm} Ambedkar Dukkipati \\
Department of Computer Science and Automation \\
Indian Institute of Science\\
{\tt\small \{gauravp, ambedkar\}@iisc.ac.in}
}


\maketitle
\ifwacvfinal\thispagestyle{empty}\fi

\begin{abstract}
\noindent
Deep convolutional networks have achieved the state-of-the-art for semantic image segmentation tasks. However, training these networks requires access to densely labeled images, which are known to be very expensive to obtain. On the other hand, the web provides an almost unlimited source of images annotated at the image level. How can one utilize this much larger weakly annotated set for tasks that require dense labeling? Prior work often relied on localization cues, such as saliency maps, objectness priors, bounding boxes etc., to address this challenging problem. In this paper, we propose a model that generates auxiliary labels for each image, while simultaneously forcing the output of the CNN to satisfy the mean-field constraints imposed by a conditional random field. We show that one can enforce the CRF constraints by forcing the distribution at each pixel to be close to the distribution of its neighbors. This is in stark contrast with methods that compute a recursive expansion of the mean-field distribution using a recurrent architecture and train the resultant distribution. Instead, the proposed model adds an extra loss term to the output of the CNN, and hence, is faster than recursive implementations. We achieve the state-of-the-art for weakly supervised semantic image segmentation on VOC 2012 dataset, assuming no manually labeled pixel level information is available. Furthermore, the incorporation of conditional random fields in CNN incurs little extra time during training.\end{abstract}

\section{Introduction}
Semantic image segmentation is the problem of assigning the pixels of an image to a selected set of predefined labels, based on the semantic structure that the pixel belongs to. Most successful models for semantic image segmentation employ a variation of CNN for computing the probability distribution over the classes for each pixel. During inference, these distributions are fed as unary potentials to a fully connected CRF with Gaussian edge potentials, and a joint labeling for the pixels of the image is inferred from the CRF. The work by Kr\"{a}henb\"{u}hl \& Koltun~\cite{koltun2011efficient}, allows for efficient inference in such models.

Successful semantic image segmentation requires access to a large number of images that have been densely labelled. However, dense labeling of images is an expensive and time-consuming operation~\cite{papandreou2015weakly,pinheiro2015image,pathak2014fully}.  Therefore, the number of densely labeled images available is usually a minuscule percentage of the total set of images. 
Hence, the models that rely solely on densely labeled images, are limited in their scope. These models will be referred to as fully supervised models in the sequel.

The limitations of fully supervised models has necessitated the development of models that can incorporate weakly labeled images for training. These include models that utilize bounding box prior~\cite{lempitsky2009image,dai2015boxsup,papandreou2015weakly,Xu_2015_CVPR}, few points per class~\cite{bearman2016s} and image-level labels~\cite{vezhnevets2011weakly,papandreou2015weakly,pinheiro2015image}. Of particular interest are models that rely on image-level labels only, since the web provides an almost unlimited source of weakly annotated images. 

Unfortunately, the decoupled CNN-CRF combination (or CNN alone) fares poorly, when only image-level labels are available~\cite{pinheiro2015image,pathak2015constrained}. To alleviate this problem, several researchers have resorted to the use of localization cues, such as saliency and attentions maps~\cite{wei2016stc,kolesnikov2016seed} or objectness priors~\cite{pinheiro2015image,wei2016learning}, thereby improving performance to an extent. Improvements in CNN architecture for segmentation~\cite{chen2016deeplab,yu2015multi}, have further aided in improved performance.

In this paper, we propose a model that learns to output segmentation masks using only image-level labels without the aid of localization cues or saliency masks. In particular, we enforce a pixel-label loss as well as a neighborhood loss on the output of a CNN. Since real pixel-labels are unavailable, we map the output of the CNN to auxiliary pixel labels to get an approximate segmentation mask. The neighborhood loss allows us to enforce the constraints imposed by conditional random field on the output of the CNN thereby forcing it to generate crisp segmentation masks that align with the boundary of the object
\subsection*{Contributions} 
Our contributions are as follows. 
\textbf{(1)} We propose a new interpretable model for weakly supervised semantic segmentation.
\textbf{(2)} The model is trained by imposing pixel-label and neighborhood loss functions on the output of a fully convolutional neural network.
\textbf{(3)} We prove that by imposing neighborhood loss on the output of the CNN, the output of the CNN is forced to satisfy the constraints imposed by a conditional random field.
\textbf{(4)} We achieve an accuracy of 52.01 \% on the \emph{test set} and 51.6 \% on the validation set of Pascal VOC-2012, which is the state of the art for methods that do not employ any pixel-level labels.



\section{Preliminaries and background}
\subsection*{Notations}
The probability distributions are indicated by lower case letters, for example, $p$ and $q$. 
Subscripts in the distribution indicate the location, whereas superscripts indicate the name of the distribution. 
The label at $i^{th}$ location is denoted as $z_i$, while any distribution at the corresponding location is denoted as $p_i$ with an appropriate superscript indicating the type of distribution. 
The labels in the segmentation mask form a grid, which is denoted by $\mathcal{G}$. The entire segmentation mask is denoted as $\fatzs$.

\subsection*{Conditional random fields for semantic image segmentation} \label{sec:CRF}
A conditional random field is an example of undirected graphical model that models the conditional distribution of output given the input, when the output is structured in the form of factors. For the problem of semantic image segmentation, the image is the input, whereas the pixel-level labels form the output. A conditional random field is completely characterized by its potential functions.

Traditionally, conditional random fields employ two forms of potential functions: unary potentials and binary potentials. A unary potential $\phi_i$ is a function of the conditioning variable $\fatxs$ and a single output variable $z_i$. The unary potentials encode the suitability of assigning a specific label at a specific location. A popular approach is to learn a local classifier for each location in the image, using the features extracted locally from the image. The unary potential for a specific label at a specific location is then equated to the negative log-probability of observing the label at that location. These local classifiers have largely been replaced by convolutional neural networks that consider local as well as global information~\cite{zheng2015conditional,chen2016deeplab}.

The binary potential $\phi_{ij}(z_i, z_j, \fatxs)$ measures the compatibility of two labels $z_i$ and $z_j$ for locations $i$ and $j$. 
The most commonly used binary potential is the contrast sensitive Potts model~\cite{boykov2001interactive,shotton2006textonboost,kohli2009robust}, which captures the difference in color among neighboring locations, that is
\begin{equation} 
\phi_{ij}(z_i, z_j, \fatxs) =  \alpha \exp\left(- \beta{||x_i - x_j||^2}\right) \mathbbm{1}(z_i \neq z_j)\,,
\end{equation}
where $\alpha$ and $\beta$ are non-negative scalar constants.

A CRF with unary and binary potentials only, is referred to as a \textit{pairwise CRF}. For a given input image $\fatxs$ and a segmentation mask $\fatzs = (z_i, i \in \mathcal{G})$, the joint distribution of a pairwise CRF is given below:
\begin{equation}~\label{pairwiseCRF}
\begin{aligned}
p&(\fatzs|\fatxs) \\
&= \frac{1}{\mathcal{Z}}\exp \left( -\sum_{i \in \mathcal{G}} \phi_i(z_i, \fatxs) - \sum_{i \in \mathcal{G}}\sum_{j \in \mathcal{N}(i)} \phi_{ij}(z_i, z_j, \fatxs)\right) \,,
\end{aligned}
\end{equation}
where $\mathcal{N}(i)$ indicates the neighbors of the $i^{th}$ node.

\section{Proposed Model}
\begin{figure*}
\center
\includegraphics[width=.8\linewidth, height=.35\linewidth]
                   {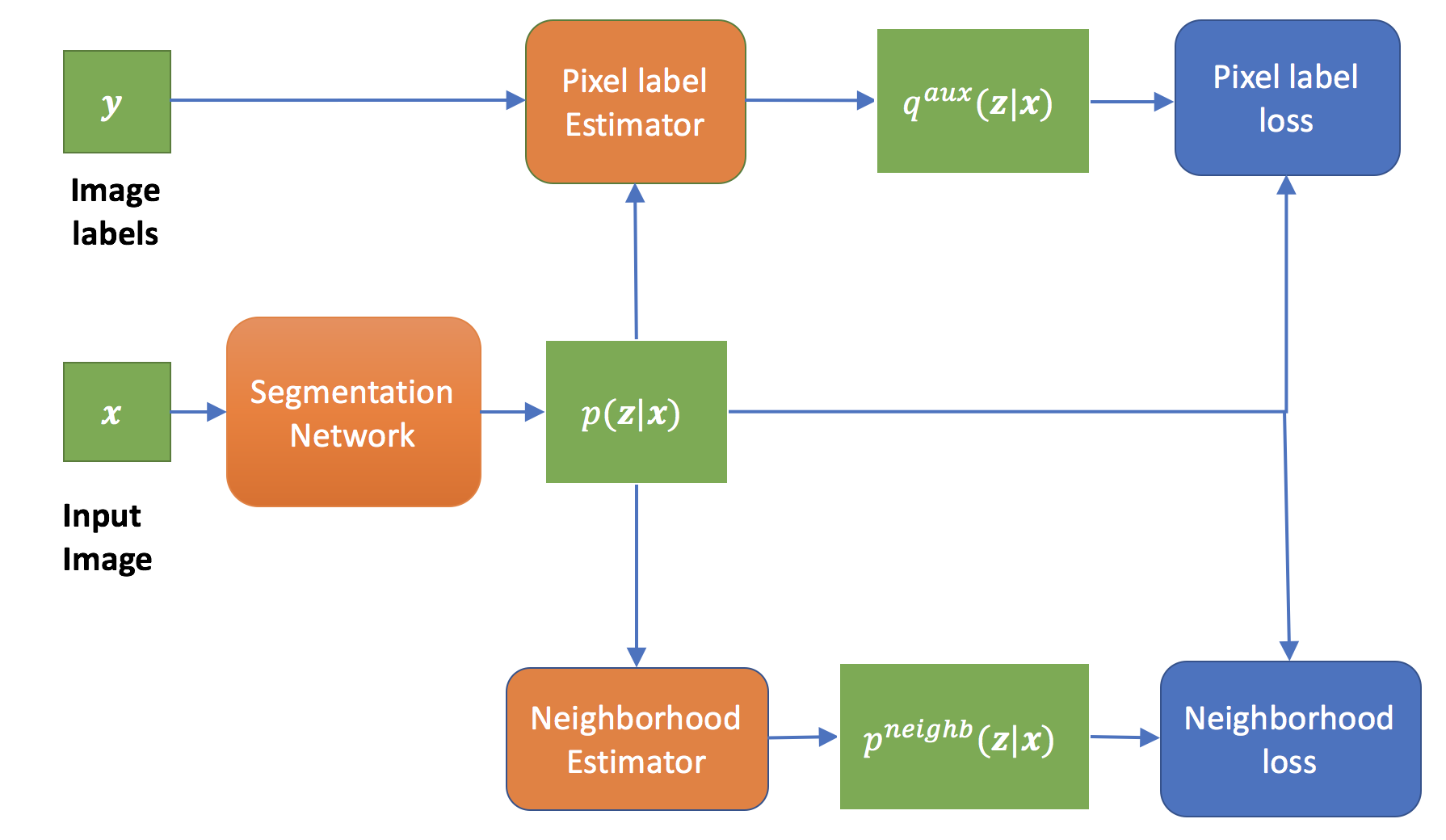}
\caption{The input image is fed through a fully convolutional network to generate a distribution over segmentation masks $p(\fatzs|\fatxs)$. The pixel label estimator incorporates the image label information in the distribution to generate $\qfake(\fatzs|\fatxs)$. We force the output of segmentation network to be close to this updated distribution. Simultaneously, the neighborhood loss enforces the output of the segmentation network to be close to the distribution computed from its neighbors.}                   
\label{networks}
\end{figure*}                

\subsection{Overview}
Given a set of images and their corresponding image-level labels, the aim is to learn a model that can output pixel-level labels from the input image. It is important to note that pixel-level labels are not provided during training, and hence, constitute the latent variables in the current model.  An image is fed through a \emph{segmentation network} that outputs a distribution over the labels for each pixel location $p(\fatzs|\fatxs)$. We refer to this distribution as the \emph{predicted distribution}, since this is the only distribution that will be required during inference. Our aim is to ensure that the predicted distribution constitutes a valid segmentation mask for the input image. Hence, we impose multiple losses on the predicted distribution. In particular, the \emph{pixel-label estimator} incorporates the image-label information in the predicted distribution to generate a distribution over pixel-level labels $\qfake$. This distribution can be thought of as an auxiliary ground truth, since the true pixel level labels are not available. The segmentation network is trained using the auxiliary ground truth. 

Next, the \emph{neighborhood estimator} computes a smooth version of the output distribution by averaging the output of the neighbors for each location. We force the output of the CNN to be close to the neighborhood distribution. We further show that this is equivalent to enforcing the constraints of a CRF on the output of a CNN. 

\begin{figure*}
\center
\includegraphics[width=.8\linewidth, height = .35\linewidth]
                   {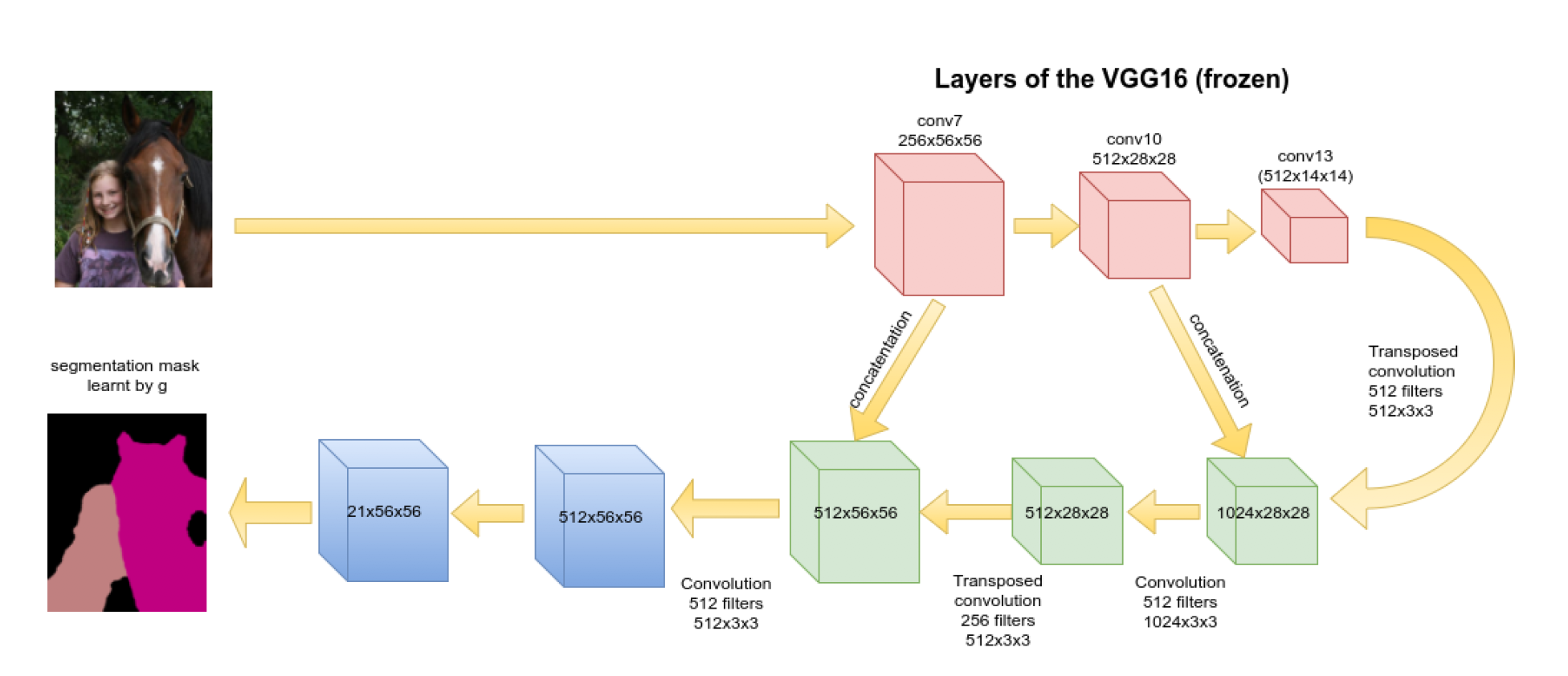}
\caption{The segmentation network in the proposed model. Except for the last layer, each convolutional layer in the inference network is followed by a ReLU layer and a batch normalization layer. The last layer is followed by exponentiation and normalization to get $\phi$.}                   
\label{fig:segmentation_network}
\end{figure*}                
In sequel,  $\fatxs$ represent the input image and $\fatzs$ represents the segmentation mask. The label of the pixel at location indexed by $i$ is denoted as $z_i$. The various components involved in the model are discussed below.
\subsection{Segmentation Network}
The segmentation network receives the image as input and generates a distribution over the segmentation masks as output. As is common in CNN-based training for segmentation~\cite{chen2014semantic,long2015fully}, we assume that the output distribution over pixel-level labels factorizes completely for each location.
In particular, let $p(\fatzs|\fatxs)$ be the conditional distribution over the pixel-level labels given the image. We assume that $p(\fatzs|\fatxs) = \prod_{j \in \mathcal{G}} p_j(z_j|\fatxs)$, where $p_j$ is the distribution at location $j$, and $z_j$ is the corresponding label.   Furthermore, we assume that the distribution is parametrized by a CNN $f$, that is,
\begin{equation}
p_j(\ell|\fatxs) =  \frac{\exp(f_{j\ell}(\fatxs))}{\sum_{\ell'=0}^L \exp(f_{j\ell'}(\fatxs)} \,.
\end{equation}
The segmentation network used in this paper is shown in Figure~\ref{fig:segmentation_network}.

\subsection{Pixel Label Estimator}\label{transfer_seg}
Since ground truth information for the pixels is not available, we attempt to generate auxiliary ground truth information for each pixel from the output of the network. In particular, we infer a distribution over the labels $\qfake(\fatzs|\fatxs, \fatys)$ for each location in the image from the output distribution of segmentation network. Given this auxiliary ground truth, the classification objective can be rewritten as 
\begin{equation}
\lclass = -\mathbb{E}_{\qfake(\fatzs|\fatxs, \fatys)} \log p(\fatzs|\fatxs)\,.
\end{equation}

In absence of any restriction, the model can choose to assign all the pixels to a single class, for instance, the background class. In order to prevent this from happening, one needs to ensure that for each class present in the image, at least a certain percentage of pixels are allotted to that class. Furthermore, one also needs to ensure that no pixels are allotted to classes absent from the image. Hence, we couple the distribution with a prior to obtain the distribution over the pixel-labels. 
\begin{equation}
\qfake_j(\ell|\fatxs, \fatys) = \frac{p_j(\ell|\fatxs)\prior(\ell|\fatys)}{\sum_{\ell'}p_j(\ell'|\fatxs)\prior(\ell'|\fatys)}\, ,
\end{equation}
for $\ell = 0, \ldots, L$, and $j \in \mathcal{G}$.
In order to complete the description, we define the prior distribution $\prior(\ell|\fatys)$ as below:
\begin{equation}
\begin{aligned}
\prior(\ell|\fatys) = \begin{cases}
\beta_\ell, & \text{if } \ell \in \fatys \,,\\
0, & \text{otherwise}
\end{cases}
\end{aligned}
\end{equation} 

Images in the ImageNet dataset are assumed to contain only one foreground object, and hence $\#\fatys=2$.
We learn the constants $\beta_\ell, \ell \in \fatys$ independently for each image. In particular, for each image, we learn the most \textit{non-informative} prior that can guarantee the assignment of a certain percentage of pixels to each class present in the image, while assigning no pixels to classes absent from the image. In order to quantify \textit{information}, we maximize the \textit{entropy} of the prior distribution, while simultaneously forcing it to satisfy a set of constraints. That is,
\begin{equation}
\begin{aligned}
& \underset{\beta_\ell, \ell \in \fatys}{\text{minimize}} & & \sum_{\ell \in \fatys} \beta_\ell \log \beta_\ell\\
& \text{subject to}
& & \frac{1}{\#\mathcal{G}}\sum_{j\in \mathcal{G}} \qfake_j(\ell|\fatxs, \fatys) \ge c_\ell, \; \ell \in \fatys \,,\\
& \text{  and} & & \sum_{\ell \in \fatys} \beta_\ell = 1, \text{ and } \beta_\ell \ge 0, \; \ell \in \fatys.
\end{aligned}
\end{equation}
The constants $c_\ell$ dictates the percentage of pixels that are guaranteed to belong to class $\ell$, if label $\ell$ is present in the image. We choose $c_\ell=.4$ for the background class, and $c_\ell=.2$ for all the other object classes present in the image.

For images in the ImageNet dataset, $\#\fatys=2$, and hence the above optimization problem contains only two variables, $\beta_0$ and $\beta_\ell$, where $\ell$ is the label of the foreground object is present in the image. Furthermore, by equating $\beta_0$ to $1-\beta_\ell$, we further reduce the number of variables from $2$ to $1$.
Hence, the above optimization problem reduces to a constrained optimization in a single variable which can be solved very efficiently. This approach is discussed in further detail in the Appendix.

\subsection{Neighborhood Estimator}\label{neighborhood_seg}
To ensure the correct alignment between the predicted boundaries and the actual boundaries, we utilize the following information: Pixels that lie close together and have similar color, also have the same label. 
Hence, we force the distribution at location $i$ to be close to the distribution of its neighbors. Towards that end, we compute a neighborhood distribution for each location $\pnei$, and minimize the KL-divergence between the output distribution at that location and the corresponding neighborhood distribution. The corresponding objective is given by
\begin{equation}
\lneigh = -\KL(p(\fatzs|\fatxs)||\pnei(\fatzs|\fatxs)))\,.
\end{equation}
The combined objective is given by
\begin{equation}
\mathcal{L} = \lclass + \lambda\lneigh\,,
\end{equation}
for some constant $\lambda \ge 0$. 
We propose two approaches for computing the neighborhood distribution.

\noindent\textbf{Weighted mean:}
In this approach, the neighborhood distribution is computed as follows:
\begin{equation}
\pnei_i(\ell|\fatxs) = \frac{\sum_{j\neq i} k(f_i(\fatxs),f_j(\fatxs)) p_j(\ell|\fatxs)}{\sum_{j\neq i}k(f_i(\fatxs),f_j(\fatxs))}\,,
\end{equation}
for $\ell \in \{0, \ldots, L\}$. Here, $k(f_i(\fatxs),f_j(\fatxs))$ is a measure of similarity between the locations $i$ and $j$. For our purpose, we define the neighbors as all the locations that lie close to the current location, and the corresponding pixels have similar color. Hence, we use the contrast sensitive two-kernel potential~\cite{koltun2011efficient} defined in terms of pixel locations $i$ and pixel brightness $x_i$ as follows:
\begin{equation}
\begin{aligned}
k(f_i(\fatxs),f_j(\fatxs)) &= k_1(f_i(\fatxs),f_j(\fatxs)) + k_2(f_i(\fatxs),f_j(\fatxs)) \\
=& w_1 \exp\left(- \frac{|x_i-x_j|^2}{2 \theta_\alpha^2} - \frac{|i-j|^2}{2 \theta_\beta^2}\right) \\  
& + w_2 \exp\left(- \frac{|i-j|^2}{2 \theta_\gamma^2} \right)\,,
\end{aligned}
\end{equation}
where, $f_i(\fatxs) = [x_i, i]$. Here, $w_1, w_2, \theta_\alpha, \theta_\beta$ and $\theta_\gamma$ are hyperparameters that are fixed during training. As discussed in~\cite{koltun2011efficient}, the second term prevents the formation of small isolated regions as segments.

\noindent\textbf{Exponentiated weighted mean:}
As the name suggests, the neighborhood distribution in this approach is obtained by exponentiating the weighted mean. The exponentiation causes the neighborhood distribution to be sharper, resulting in high confidence predictions.
\begin{equation}~\label{neighb}
\pnei_i(\ell|\fatxs) = \frac{1}{\mathcal{Z}_i}\exp\left(\sum_{j\neq i}\hat{k}_\fatxs(i, j)p_j(\ell|\fatxs)\right)\,,
\end{equation}
where $$\hat{k}(f_i(\fatxs),f_j(\fatxs)) = \frac{k(f_i(\fatxs),f_j(\fatxs))}{\sum_{j' \neq i} k(f_i(\fatxs),f_{j'}(\fatxs))}$$ and $\mathcal{Z}_i$ is a normalization constant that ensures that the above distribution sums up to $1$. 
Next, we show the connection between the exponentiated weighted mean and CRF.
\subsection{Connections with CRF}
In this section, we provide a formal justification for the choice of the neighborhood-based objective function. In particular, we will show that the objective emerges naturally, when a CRF is used as a prior while computing the conditional log-likelihood.

Given an image $\fatxs$, let the CRF prior over the segmentation masks be defined as below:
\begin{equation}
p^{\text{CRF}}(\fatzs|\fatxs) = \frac{1}{\mathcal{Z}}{\exp\left(-\sum_{i<j}\phi_{ij}(z_i, z_j, \fatxs)\right)}\,,
\end{equation}
where $\mathcal{Z}$ is the normalization constant. Note that the prior distribution has no unary potentials. We will further assume that binary potentials $\phi_{ij}$ have no trainable parameters.

The prior provides a distribution over all possible segmentation masks for a given image. Furthermore, let $\psi_{ij}(z_i, z_j, \fatxs)$ has the form
\begin{equation}
\psi_{ij}(z_i, z_j, \fatxs) = -k(f_i(\fatxs),f_j(\fatxs)) \mathbbm{1}(z_i=z_j) \,,
\end{equation} 
for some choice of kernel $k$. The corresponding CRF prior gives low probability to masks $\fatzs$ that assign different labels to pixels $i$ and $j$ with high similarity (that is, high $k(f_i(\fatxs),f_j(\fatxs))$). This is a reasonable prior assumption about the segmentation mask of an image. Note that the prior does not penalize masks that assign the same label to pixels $i$ and $j$ with low similarity. This allows the inclusion of object classes with multicolored instances. For instance, the dress a person wears will often be colored differently from his skin color.

If the CRF prior is approximated by a fully factorized distribution, the resultant distribution will have to satisfy the constraints entailed in Proposition~\ref{mean_CRF}.
\begin{prop}~\label{mean_CRF}
Let $\qmf$ be the mean field approximation to the CRF prior $\pcrf$. That is, among all distributions of the form $Q(\fatzs) = \prod_{i \in \mathcal{G}} Q_i(z_i)$, let $\qmf$ be the one that minimizes $\KL(Q||\pcrf)$. Then the distribution $\qmf$ satisfies the following constraints:
\begin{equation}
\qmf_i(\ell) = \frac{1}{\mathcal{Z}_i}\exp\left(\sum_{j\neq i}k(f_i(\fatxs),f_j(\fatxs))\qmf_j(\ell)\right)\,,
\end{equation}
for $\ell \in \{0. \ldots, L\}$ and $i \in \mathcal{G}$.
\end{prop}

These constraints are referred to as mean field constraints. The proof of the Proposition is given in the Appendix. $\mathcal{Z}_i$ is the normalization constant that ensures that the distribution sums up to $1$.

Coming back to the predictive distribution for segmentation masks in our model $p$, if one wishes to impose a CRF prior on $p$, one must force it to satisfy the mean-field constraints, that is,
\begin{equation}
p_i(\ell) = \frac{1}{\mathcal{Z}_i}\exp\left(\sum_{j\neq i}k(f_i(\fatxs),f_j(\fatxs))p_j(\ell)\right)\,,
\end{equation}
for $\ell \in \{0. \ldots, L\}$ and $i \in \mathcal{G}$.
The distribution on the RHS of the above equation is exactly the neighborhood distribution $\pnei$ of equation~\eqref{neighb}, with the exception that the kernel $k$ is normalized in~\eqref{neighb}. The distribution $p$ is defined in terms of the output of a neural network. Hence, instead of the equality constraints imposed by the mean-field,  we add the term $\KL(p||\pnei)$ to the objective. The KL-divergence term forces the output of the network to satisfy the mean field constraints imposed by the CRF prior.

\noindent\textbf{Note:} The binary potential used in the CRF prior in this section is given by $\phi_{ij}(z_i, z_j, \fatxs) = -k(f_i(\fatxs),f_j(\fatxs))\mathbbm{1}(z_i=z_j)$. In contrast, the binary potential commonly used for semantic segmentation has the form $\phi(z_i, z_j, \fatxs) = k(f_i(\fatxs),f_j(\fatxs))\mathbbm{1}(z_i\neq z_j)$. However, when the kernel $k$ is normalized at the pixel-level (as has been suggested in~\cite{koltun2011efficient}), the resultant distributions are exactly the same.

\section{Relation with similar works}
Recent works on semantic segmentation using deep architectures have focused on pairwise CRFs with only unary and binary potentials. The unary potentials were specified by the output of a CNN while the binary potentials have no learnable parameters~\cite{chen2016deeplab,zheng2015conditional,schwing2015fully}. The work in~\cite{lin2016efficient} allows the binary potentials to be learnable as well.

Our work significantly differs from the above mentioned works in the learning algorithm used for training the parameters of the CRF. Most of the works that combine CNNs with CRFs use piecewise learning~\cite{sutton2005piecewise,lin2016efficient,chen2016deeplab}, that is, the energy function is decomposed into its potentials, and each potential is normalized individually. For instance, if $\phi_1, \ldots, \phi_K$ are the potentials that form the energy function, the piecewise approximation to the objective is given by
\begin{equation}
\log \prod_{k=1}^K \frac{\exp{\phi_k(\fatxs)}}{\mathcal{Z}_k} \,,
\end{equation}
where $\mathcal{Z}_k$ is the normalization constant for the $k^{th}$ potential. 

Hence, in a pairwise CRF, with unary potentials given by the output of a CNN, each output location of the CNN is trained independently. Furthermore, the contribution of the pairwise potentials is not incorporated during training of the parameters of the CNN. Hence, the training is equivalent to the training of several independent classifiers, one for each location in the segmentation mask. 

While piecewise training is extremely efficient, the lower bound that it optimizes is a very weak lower bound of true log-likelihood. By training the local classifier without incorporating the binary potentials, we ignore the dependence among labels of nearby pixels with similar color. More importantly, it is completely unsuitable when the pixel-level labels are absent, which is the main concern of this paper. 

More recently, several authors have considered training the mean field approximation $q(\fatzs)$ rather than the actual CRF distribution $p(\fatzs|\fatxs)$~\cite{kraehenbuehl2013parameter,zheng2015conditional,schwing2015fully} for semantic segmentation. The mean-field approximation $q(\fatzs)$ for a distribution $p(\fatzs)$ is the distribution that minimizes the KL-divergence $\KL(q(\fatzs)||p(\fatzs))$. By computing the gradient of the KL-divergence with respect to $q$, and equating it to $0$, one can obtain an iterative algorithm for finding the minima. For a pairwise CRF, the mean-field update equations are given by
\begin{align*}
& q^{(t+1)}_i(z_i) \\ 
&= \frac{1}{\mathcal{Z}_i}\exp\left(-\phi_i(z_i, \fatxs) - \sum_{j \in \mathcal{N}(i)} \mathbb{E}_{z_j \sim q^{(t)}_j}\phi_{ij}(z_i, z_j, \fatxs) \right) \,,
\end{align*} 
where $q^{(t)}_i$ is the distribution of the $i^{th}$ location of the mean field approximation at the $t^{th}$ iteration.

Hence, the mean field approximation at the $(t+1)^{st}$ iteration can be defined recursively, as a function of the mean-field approximation at the $t^{th}$ iteration and the potential functions. Consequently, the gradient of the mean-field distribution for $(t+1)^{st}$ iteration can be written as a function of the gradient of the approximation at the $t^{th}$ iteration and the gradient of the potential functions. This approach for training CRFs has been used in~\cite{kraehenbuehl2013parameter,zheng2015conditional,schwing2015fully}.


\section{Experimental setup}\label{setup_seg}
\subsection{Network architecture}
For our experiments, we have used pretrained VGG16 network  (trained on ImageNet for classification; torchvision\footnote{https://github.com/pytorch/vision/tree/master/torchvision}), and we have modified it for the task of semantic segmentation. The VGG16 network consists of 13 convolutional layers and 3 fully connected layers. First, we removed the fully connected layers and the last pooling layer. The resultant network receives images of size $224\x224$, and generates $512$ feature maps  of size $14\x14$. The receptive field of a neuron in the last convolutional layer is $196\x 196$, and hence, it nearly encompasses the entire input image. This implies that every neuron in the last layer has access to almost the entire image. This network serves as an encoder in our model.

To learn fine-grained contours of objects, we added skip connections from the layers of the encoder to the layers of the decoder. The receptive-field size of the neurons in the encoder is much smaller than their counterpart in the decoder. Hence, they have access to more fine-grained information. We performed $1\x1$ convolution to the output of the layers of the encoder before concatenating them with the layers of the decoder. The concatenated output is fed through multiple layers of convolution and non-linearities, to allow the final prediction layer to learn non-linear mapping of the local and global information. The final architecture of our model is shown in Figure~\ref{networks}.

\subsection{Dataset}
Models that use weak-supervision for training often employ a much larger dataset than fully-supervised models. For instance, the model in~\cite{wei2016stc} mines the pages of \textit{Flickr}\footnote{https://www.flickr.com/}, to generate the training data. Similarly, the model in~\cite{saleh2016built} uses a subset of Flickr dataset~\cite{huiskes2010new} for training. A clean subset of the ImageNet dataset is used in~\cite{hou2016mining}, while a much larger subset of the same dataset is used in~\cite{pinheiro2015image}. More importantly, almost all the approaches for semantic segmentation utilize the ImageNet dataset for pretraining the network.

We follow the experimental setup of~\cite{pinheiro2015image}. In particular, we downloaded images of objects belonging to the $20$ object classes in the VOC 2012 dataset~\cite{pascal-voc-2012} from the Imagenet database~\cite{ILSVRC15}. Several authors mine the dataset further to obtain a set of simple images which contain the object against a plain background. However, we choose to use the entire dataset to minimize manual dependence. A script to download the ImageNet classes used in training, will soon be available.

For testing, we use the validation and test set of VOC 2012 dataset. For most of our experiments, we use the validation set only, since the ground truth is publicly available. The test set is used only for comparing the final model against the state-of-the-art.

Almost all approaches for weakly supervised semantic segmentation report the results on VOC 2012 dataset. This is primarily because the classes in VOC 2012 are discrete object categories such as sheep, person, dog, etc. 
Most images in VOC 2012 contain very few of these object classes, and hence, it is easy to utilize weak labels for training. Two classes that almost always occur together, will be impossible to discern using weak labels only, for instance, road and sky. 

\subsection{Training protocol}
Stochastic gradient descent (SGD) with a minibatch of $20$ images is used for training. The initial learning rate for the pretrained layers is set at $0.001$, while the initial learning rate for newly added layers is set at $.01$. The momentum and the weight decay for gradient descent is set at $0.9$ and $.00005$ respectively. We halve the learning rate after every $4000$ iterations. All the networks are trained for $40000$ iterations.

\section{Experiments}
We use a weighted combination of the objective based on auxiliary labels and the neighborhood based objective in our experiments. For our comparisons, we have limited ourselves to models that do not employ pixel level information. Hence, we have excluded the models that employ agnostic segmentation or saliency masks that have been manually labelled~\cite{hou2016mining,wei2017object} with the exception of STC~\cite{wei2016stc}.

\begin{equation}
\mathcal{L} = \mathbb{E}_{\qfake(\fatzs|\fatxs, \fatys)} \log p(\fatzs|\fatxs) - \lambda\KL(p(\fatzs|\fatxs)||\pnei(\fatzs|\fatxs))
\end{equation}
The model is trained on the ImageNet subset and evaluated on the VOC 2012 validation set.

\noindent\textbf{Hyperparameters:} We use the same hyperparameter settings for the kernel as used in the publicly available code associated with the paper~\cite{koltun2011efficient}. In particular, we modify the code in~\cite{koltun2011efficient} to allow us to compute the weighted and exponentiated weighted mean. To account for the fact that our network reduces the segmentation mask to one-fourth of the input image, we divide $\theta_\beta$ and $\theta_\gamma$ by $4$. during training.

\begin{figure}
\centering

\includegraphics[width=.4\textwidth]{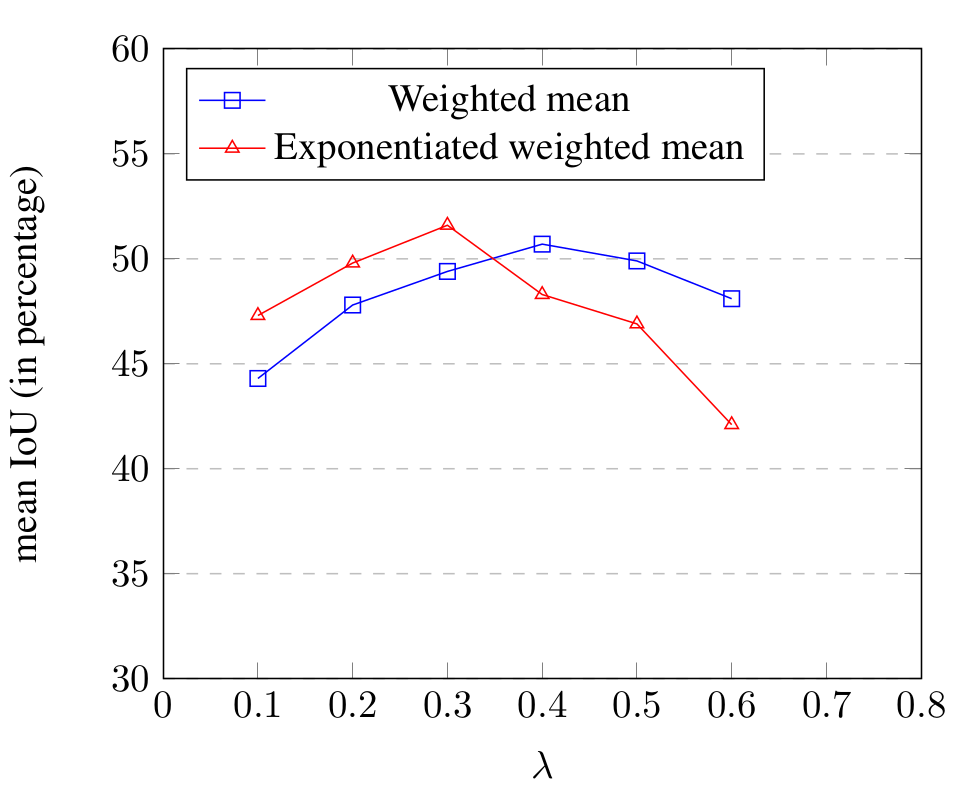}
\caption{When weighted mean is used as neighborhood distribution}
  \label{lambdaplot1}
\end{figure}

We also evaluate the effect of the hyperparameter $\lambda$ on the performance of the model. Note that the KL-divergence term will be minimized when all the pixels are assigned the same label. Hence, when higher weight is given to the KL-divergence term, all the pixels get assigned to the background class (since it is the most prominent class). In contrast if $\lambda$ is close to $0$, the boundaries of segmentation masks learnt by the model do not coincide with the boundaries of the objects. We vary the value of $\lambda$ from $0$ to $1$ to study the effects of $\lambda$ on the performance of the model. The results are shown in Figures~\ref{lambdaplot1}. As can be observed, when weighted mean is used as the neighborhood distribution, the model remains quite stable to the choice of $\lambda$. However, the overall performance achieved is worse as compared to the model that relies on exponentiated weighted mean (51.6\% vs 50.7\%). 
The segmentation masks generated by using exponentiated weighted mean at $\lambda=.3$ are shown in Figure~\ref{masks}.

We also compare the proposed model against other models for weakly supervised segmentation on the VOC 2012 val set. Note that generating segmentation masks involves identifying the object and identifying the boundary of the object. If the boundary of the object can be identified correctly, segmentation involves classifying the segmented object which can be done using a classifier with relative ease. Hence, models that employ networks pretrained on boundary detection are obviously at an advantage and hence, excluded from comparison. This includes~\cite{hou2016mining} which employs groundtruth saliency masks, since the boundaries in the saliency masks used in this paper, coincide with the boundaries of the segmentation masks.

The models used for comparison are (i) MIL+ILP~\cite{pinheiro2015image}. (ii) EM-Adapt~\cite{papandreou2015weakly}, (iii) CCNN~\cite{pathak2015constrained}, (iv) SEC~\cite{kolesnikov2016seed} and (v) STC~\cite{wei2016stc}. Among these models, SEC employs separate convolution networks for computing the saliency maps for foreground and background. On the other hand, the saliency network in STC is trained using dense (pixel-level) supervision. The results are given in Table~\ref{validation}. Note that the proposed model is similar in spirit to CCNN~\cite{pathak2015constrained} and EM-Adapt~\cite{papandreou2015weakly}, with the exception of the incorporation of a neighborhood based objective. One can observe that the incorporation of neighborhood information results in drastically improved performance.

Finally, we evaluate the performance of our model on the VOC 2012 test set by submitting our results to the evaluation server of VOC 2012. An anonymous view of the results is available at \textit{http://host.robots.ox.ac.uk:8080/anonymous/BEC3EB.html}. We achieve an accuracy of $52.02\%$ which is the state-of-the-art for weakly supervised semantic image segmentation on this dataset.

\begin{table}[!ht]
\begin{center}
\begin{tabular}{|l||c|c|c|c|c|c|}
\hline
class & \begin{turn}{90}MIL+ILP\end{turn} & \begin{turn}{90}EM-Adapt\end{turn} &  \begin{turn}{90}CCNN\end{turn} & \begin{turn}{90}SEC\end{turn} & \begin{turn}{90}STC\end{turn} & \begin{turn}{90}Ours\end{turn} \\
\hline\hline
background &77.2 & 67.2 & 68.5 & 82.4 & 84.5 & \textbf{85.4}  \\
aeroplane &37.3 & 29.2 & 25.5 & 62.9 & {68.0} & \textbf{68.4} \\
bike & 18.4 & 17.6 & 17.0 & {26.4} & 19.5 & \textbf{28.3}\\
bird & 25.4 & 28.6 & 25.4 & 61.6 &60.5 & \textbf{63.6}\\
boat & 28.2 & 22.2 & 20.2 & 27.6 & {42.5} & \textbf{42.9}\\
bottle & 31.9 & 29.6 & 26.3 & 38.1 &44.8 & \textbf{54.4}\\
bus & 41.6 & 47.0 & 46.8 & 66.6 & \textbf{68.4} & 62.7\\
car & 48.1 & 44.0 & 47.1 & 62.7 & \textbf{64.0} & {62.9}\\
cat & 50.7 & 44.2 & 48.0 & \textbf{75.2} & 64.8 & 67.5\\
chair & 12.7 & 14.6 & 15.8 & \textbf{22.1}& 14.5 & 10.6\\
cow & 53.5 & 45.7 & 35.1 & \textbf{53.5} & 52.0 & 46.3\\
diningtable & 14.6 & 24.9 & 21.0 & {28.3} & 22.8 & \textbf{37.2}\\
dog & 50.9 & 41.0 & 44.5 & \textbf{65.8} &58.0& 48.7\\
horse & 44.1 & 34.8 & 34.5 & \textbf{57.8} & 55.3 & {53.8}\\
motorbike & 39.2 & 41.6 & 46.2 & \textbf{62.3}& 57.8 & 57.3\\
person & 37.9 & 32.1 & 40.7 & {62.3}& 60.5 & \textbf{64.7}\\
plant & 28.3 & 30.4 & 24.8 & 32.5 & {40.6} & \textbf{44.3}\\
sheep & 44.0 & 36.3 & 37.4 & \textbf{62.6} & 56.7 &{58.2}\\
sofa & 19.6 & 24.0 & 22.2 & 32.1 & 23.0 & \textbf{35.8}\\
train & 37.6 & 38.1 & 38.8 & {45.4} & \textbf{57.1} & 43.6\\
tvmonitor & 35.0 & 31.6 & 36.9 & 45.3 &31.2& \textbf{47.4}\\
\hline
mean &36.6 & 33.8 & 35.3 & {50.7} & 49.8 & \textbf{51.6} \\
\hline
\end{tabular}
\end{center}
\caption{Results on PASCAL VOC 2012 (mIoU in \%) \textit{val} set for weakly supervised segmentation.}
\label{validation}
\end{table}

\begin{table}[!ht]
\center
  \begin{tabular}{lll}
	\MyIm{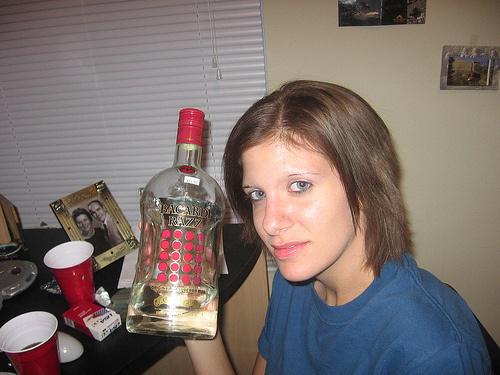} & \MyIm{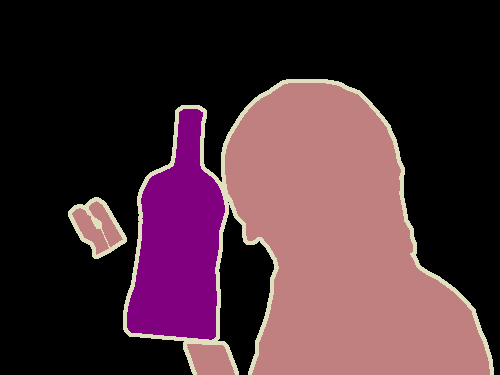} & \MyIm{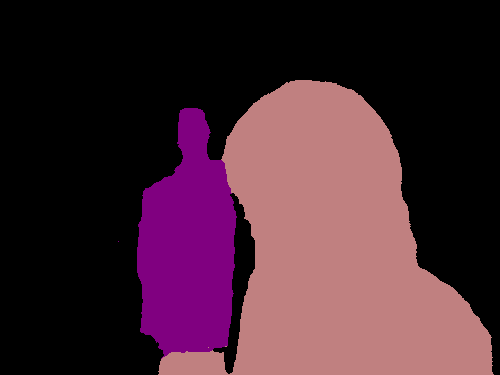} \\
	\MyIm{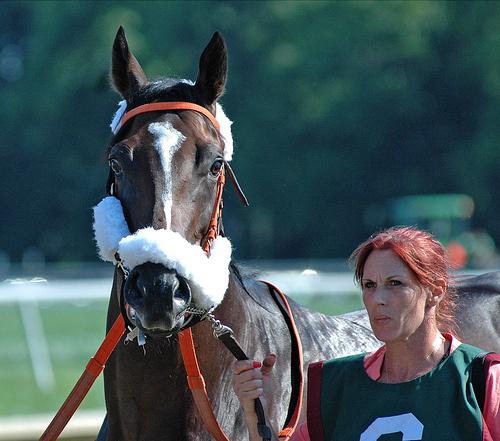} & \MyIm{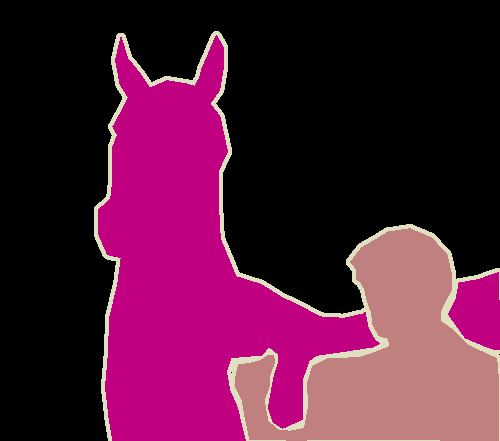} & \MyIm{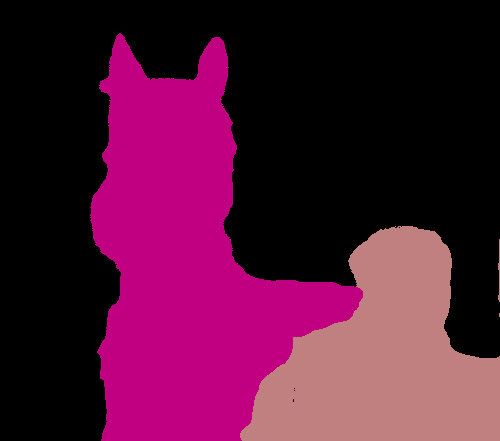} \\
	\MyIm{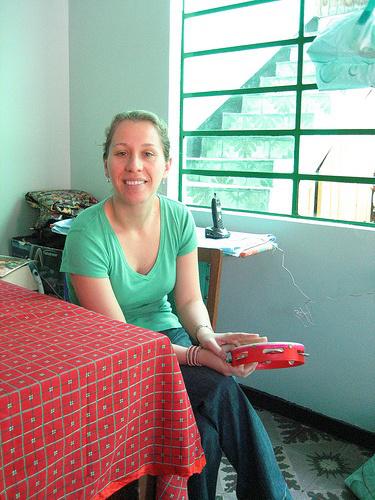} & \MyIm{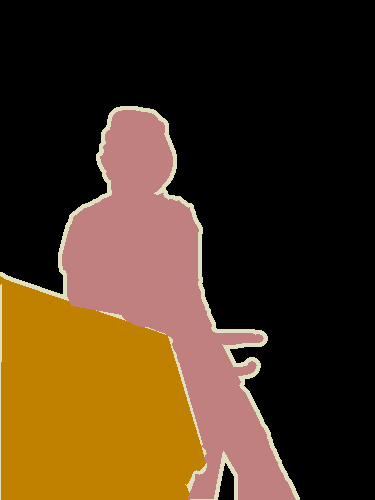} & \MyIm{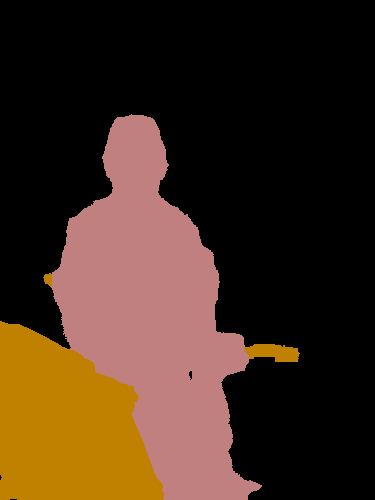} \\
	\MyIm{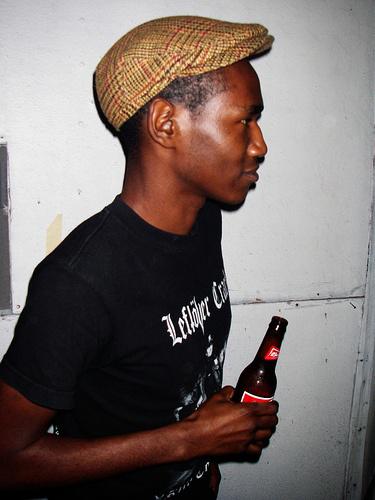} & \MyIm{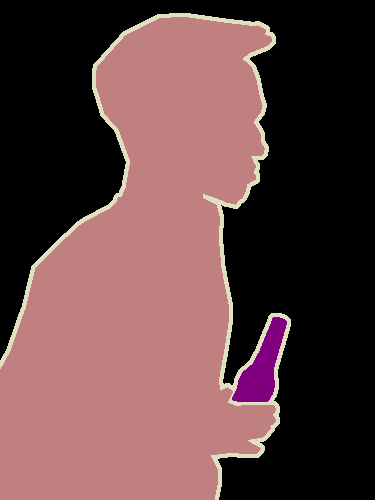} & \MyIm{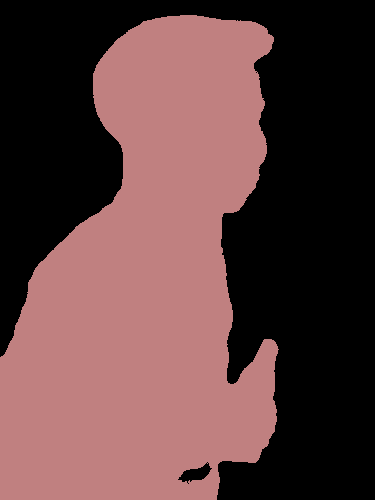} \\
	\MyIm{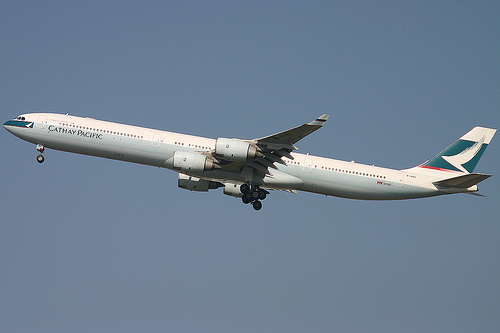} & \MyIm{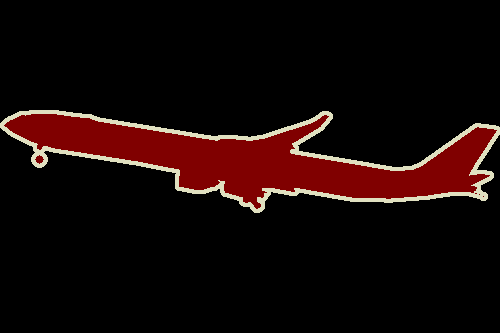} & \MyIm{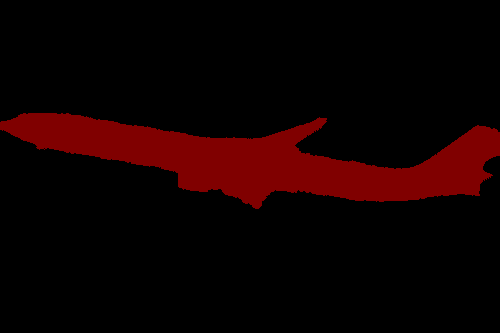} \\
  \end{tabular}
  \caption{Examples of predicted segmentation masks. The middle row is the ground truth. Note that the model has learnt to align the predicted boundaries with the true boundaries.}
  \label{masks}
\end{table}

\section{Conclusions}
In this paper, we have proposed a new model for weakly supervised semantic image segmentation that uses only image-level labels. We have shown that the output of the CNN can be forced to satisfy the constraints of a conditional random field, without explicitly evaluating the mean-field distribution at every step. We achieve this forcing the output distribution at every pixel  to be close to its neighbors. As a consequence, we achieve significant performance improvement over traditional CNNs with negligible increase in training time.

We focused on weakly annotated images in this paper, since CRFs can achieve drastic performance improvements for this task. When pixel-level information is available, forcing the pixel-level labels to be close to its neighbors will serve as an unnecessary  and often over-smooth regularizer, unless only rough object boundaries are only available. A model capable of handling rough object boundaries, can drastically reduce the time required for manually generating segmentation masks. We intend to explore the utility of the model for handling rough boundaries in the future.

\section{Acknowledgement}
Authors acknowledge financial support for the ”CyberGut”
expedition project by the Robert Bosch Centre for Cyber
Physical Systems at the Indian Institute of Science, Bengaluru.

\bibliographystyle{ieee}
\bibliography{iccv}

\clearpage
\section*{On Adaptive prior}
Let $\fatxs$ be an image and $\fatys$ be the corresponding image level labels (including the background class). The optimization problem for obtaining the pixel-labels using an adaptive prior is given below:
\begin{equation}
\begin{aligned}
& \underset{\beta_l, l \in \faty}{\text{minimize}}
& & \sum_{l \in \faty} \beta_l \log \beta_l  \\
& \text{subject to}
& & \frac{1}{m} \sum_{j=1}^m \qfake_j(l|\fatxs) \geq c_l, l \in \fatys \\
& & & \sum_{l \in \fatys} \beta_l = 1 \text{ and } \beta_l \geq 0 
\end{aligned}
\end{equation}
Here, $c_l$ is a constant that determines the minimum fraction of pixels in the image that must be assigned to class $l$.
When the images contain objects from a single object class (as is the case with ImageNet dataset), say $l$, the above optimization problem can be rewritten as shown below:
\begin{equation}
\begin{aligned}
& \underset{0 < \beta_l < 1}{\text{minimize}}
& &  \beta_l \log \beta_l + (1-\beta_l)\log (1-\beta_l) \\
& \text{subject to}
& & \frac{1}{m} \sum_{j=1}^m \qfake_j(l|\fatxs) \geq c_l \\
& & & \frac{1}{m} \sum_{j=1}^m \qfake_j(0|\fatxs) \geq c_0 
\end{aligned}
\end{equation}
where 
\begin{equation}
\qfake_j(l|\fatxs) = \frac{p(l|\fatxs)\beta_l}{p(l|\fatxs)\beta_l + p(0|\fatxs)(1-\beta_l)} \,
\end{equation}
and $\qfake_j(0|\fatxs) = 1-\qfake_j(l|\fatxs)$.

The above problem is an optimization problem in single variable. We can solve it approximately by evaluating the constraints for several values of $\beta_l \in [0,1]$. This can be achieved very efficiently on a GPU, since it involves element-wise operations on matrices of probability distributions. Let $S$ be the set of all the selected values of $\beta_l$ which satisfy the constraints. Among these values, we return the value of $\beta_l$ which minimizes the objective.

If none of the selected values of $\beta_l$ satisfy the constraints, we choose the value of $\beta_l$ which minimizes the following:
\begin{equation}
\left(c_l-\frac{1}{m} \sum_{j=1}^m \qfake_j(l|\fatxs)\right)_+ + \left(c_0-\frac{1}{m} \sum_{j=1}^m \qfake_j(0|\fatxs)\right)_+
\end{equation}

\section*{Proof of Proposition 3.1}
\textbf{Proposition 3.1}
Let $\qmf$ be the mean field approximation to the CRF prior $\pcrf$. That is, among all distributions of the form $Q(\fatxs) = \prod_{i=1}^m Q_i(x_i)$, let $\qmf$ be the one that minimizes $\KL(Q||\pcrf)$. Then the distribution $\qmf$ will have to satisfy the following constraints:
\begin{equation}
\qmf_i(z_i) = \frac{1}{\mathcal{Z}_i}\exp\left(\sum_{j\neq i}k_\fatxs(i, j)\qmf_j(z_i)\right)
\end{equation}

\begin{proof}
The proof of this result can be obtained by differentiating the KL-divergence divergence with-respect-to the components of $Q$ and equating it to $0$. Towards that end, we expand the KL-divergence term as given below:
\begin{align}
KL&(Q||\pcrf) \notag\\
=&\sum_{\fatzs} Q(\fatzs) \log {Q(\fatzs)} - \sum_{\fatzs} Q(\fatzs) \log \pcrf(\fatzs|\fatxs)\notag + C\\
=& \sum_{i=1}^m \sum_{z_i=0}^L Q_i(z_i) \log Q_i(z_i) \notag\\
&- \sum_{i<j}^m \sum_{z_i=z_j=0}^L Q_i(z_i) Q_j(z_j) k(x_i,x_j) + C
\end{align}
Here, $C$ is a constant that doesn't depend on $Q$. Differentiating the above expression with respect to $Q_i(z_i)$ and equating it to $0$, we get
\begin{equation}
Q_i(z_i) = \exp\left(1+\sum_{j\neq i} Q_j(z_j) k(x_i, x_j)\right)  
\end{equation}
Finally, since $Q_i$ is a probability distribution that sums up to $1$, we normalize $Q_i$ to get the desired result.
\end{proof}

\end{document}